%% file: output.tex
\documentclass[10pt,twocolumn,letterpaper]{article}

\usepackage[pagenumbers]{cvpr} % To force page numbers, e.g. for an arXiv version

\usepackage{graphicx} 
\usepackage{amsmath}
\usepackage{amssymb}
\usepackage{booktabs}
\usepackage{float}
\usepackage{lipsum}
\usepackage{afterpage}
\usepackage{multirow}
\usepackage[table]{xcolor}
\usepackage{colortbl}
\usepackage{hhline}
\usepackage{xcolor}
\usepackage{adjustbox}
\usepackage{tabularx}
\usepackage{siunitx}
\usepackage{pifont}
\usepackage{tikz}

\usepackage[pagebackref,breaklinks,colorlinks]{hyperref}
\usepackage[capitalize]{cleveref}
\crefname{section}{Sec.}{Secs.}
\Crefname{section}{Section}{Sections}
\Crefname{table}{Table}{Tables}
\crefname{table}{Tab.}{Tabs.}

\newsavebox{\mycustomt}
\newsavebox{\mycustomb}
\newsavebox{\mycustomr}

\definecolor{blue}{rgb}{0.92,0.96,1.0}
\definecolor{red}{rgb}{1.0,0.65,0.0}
\definecolor{brightred}{rgb}{1.0,0.0,0.0}
\definecolor{brightpurple}{rgb}{0.75, 0.0, 1.0}
\definecolor{brightgreen}{rgb}{0.0, 0.6, 0.0}
\definecolor{brightblue}{rgb}{0.0, 0.6, 1.0}
\definecolor{green}{rgb}{0.9, 0.45, 0.57}
\definecolor{yellow}{rgb}{0.97,0.98,0.86}
\definecolor{gray}{rgb}{0.5,0.5,0.5}
\definecolor{pink}{rgb}{1, 0.94, 0.96}
\definecolor{purple}{rgb}{0.97, 0.93, 1.0}
\definecolor{myblue}{rgb}{0.8, 0.8, 1}
\newcommand{\blue}{\cellcolor{blue}}
\newcommand{\yellow}{\cellcolor{yellow}}

\newcommand{\pink}{\cellcolor{pink}}
\newcommand{\purple}{\cellcolor{purple}}
\definecolor{deepblue}{rgb}{0.0, 0.0, 0.9}
\newcommand{\method}{Robo-ABC\xspace}
\sbox{\mycustomt}{%
\begin{tikzpicture}[baseline=(current bounding box.center)]
    \draw[deepblue, line width=1pt] (0,0) -- (0.3,0); % Top horizontal line
    \draw[deepblue, line width=1pt] (0.15,0) -- (0.15,-0.15); % Middle vertical line
    \draw[deepblue, line width=1pt] (0,0) -- (0,0.2); % Left vertical line
    \draw[deepblue, line width=1pt] (0.3,0) -- (0.3,0.2); % Right vertical line
\end{tikzpicture}%
}
\sbox{\mycustomb}{%
\begin{tikzpicture}[baseline=(current bounding box.center)]
    \draw[brightblue, line width=1pt] (0,0) -- (0.3,0); % Top horizontal line
    \draw[brightblue, line width=1pt] (0.15,0) -- (0.15,-0.15); % Middle vertical line
    \draw[brightblue, line width=1pt] (0,0) -- (0,0.2); % Left vertical line
    \draw[brightblue, line width=1pt] (0.3,0) -- (0.3,0.2); % Right vertical line
\end{tikzpicture}%
}
\sbox{\mycustomr}{%
\begin{tikzpicture}[baseline=(current bounding box.center)]
    \draw[brightred, line width=1pt] (0,0) -- (0.3,0); % Top horizontal line
    \draw[brightred, line width=1pt] (0.15,0) -- (0.15,-0.15); % Middle vertical line
    \draw[brightred, line width=1pt] (0,0) -- (0,0.2); % Left vertical line
    \draw[brightred, line width=1pt] (0.3,0) -- (0.3,0.2); % Right vertical line
\end{tikzpicture}%
}

\def\cvprPaperID{4845} % *** Enter the CVPR Paper ID here
\def\confName{CVPR}
\def\confYear{2024}
\begin{document}
\title{Robo-ABC: \underline{A}ffordance Generalization \underline{B}eyond \underline{C}ategories via Semantic \newline  Correspondence for Robot Manipulation}
\author{
Yuanchen Ju \textsuperscript{1,2}\thanks{Equal contribution.} \quad    
Kaizhe Hu \textsuperscript{1,2,5}\footnotemark[1] \quad 
Guowei Zhang \textsuperscript{3,2} \quad  
Gu zhang \textsuperscript{1,4,2} \\ 
Mingrun Jiang \textsuperscript{2} \quad  
Huazhe Xu \textsuperscript{2,5,1}\thanks{{Corresponding author: \href{mailto: huazhe_xu@mail.tsinghua.edu.cn}{huazhe\_xu@mail.tsinghua.edu.cn}.}}  \\
\textsuperscript{1}Shanghai Qi Zhi Institute \quad 
\textsuperscript{2}IIIS, Tsinghua University\quad 
\textsuperscript{3}School of Software, Tsinghua University\\
\textsuperscript{4}Shanghai Jiao Tong University \quad
\textsuperscript{5}Shanghai AI Lab\quad\\
\href{https://TEA-Lab.github.io/Robo-ABC}{https://TEA-Lab.github.io/Robo-ABC}
}

\twocolumn[{%
\maketitle
 \vspace{-0.5cm}
\begin{center}
    \centering
    \captionsetup{type=figure}
    % \vspace{-1em}
    \includegraphics[width=1.0\linewidth]{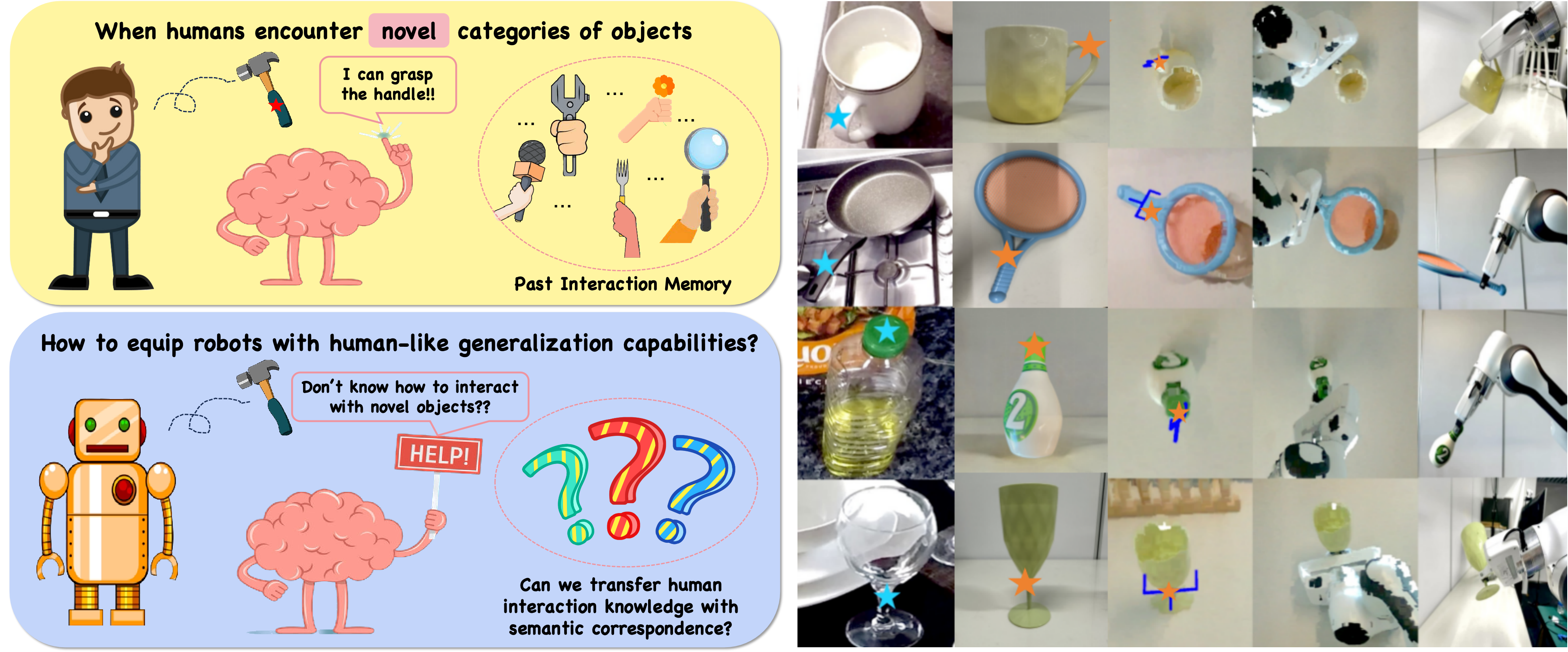}
    \captionof{figure}{\textbf{Overview.}  The \textbf{left} illustrates the key insight of Robo-ABC (\textcolor{brightred}{\Pisymbol{pzd}{72}} represents the contact point). Our goal is to endow robots with the semantic correspondence ability as humans, which can generalize the object affordance across categories in manipulation tasks. The columns on the \textbf{right} in order, are source images (\textcolor{brightblue}{\Pisymbol{pzd}{72}}  represents contact points which are extracted from human videos), corresponding attention maps on the target images  (\textcolor{orange}{\Pisymbol{pzd}{72}} represents inferred contact points on unseen objects), grasp poses (Grasp poses are represented by \ \usebox{\mycustomt}\ , which are generated according to the contact points \textcolor{orange}{\Pisymbol{pzd}{72}}), point cloud during grasping, and the final successful grasp results.} 
    \label{fig:teaser}
\end{center}
}]
\renewcommand{\thefootnote}{\fnsymbol{footnote}}
\footnotetext[1]{Equal contribution.}
\footnotetext[2]{Corresponding author: \href{mailto: huazhe_xu@mail.tsinghua.edu.cn}{huazhe\_xu@mail.tsinghua.edu.cn}.}
%%%%%%%%% ABSTRACT

\begin{abstract}
\vspace{-0.4cm}
 Enabling robotic manipulation that generalizes to out-of-distribution scenes is a crucial step toward open-world embodied intelligence. For human beings, this ability is rooted in the understanding of semantic correspondence among objects, which naturally transfers the interaction experience of familiar objects to novel ones. Although robots lack such a reservoir of interaction experience, the vast availability of human videos on the Internet may serve as a valuable resource, from which we extract an affordance memory including the contact points. Inspired by the natural way humans think, we propose \method: when confronted with unfamiliar objects that require generalization, the robot can acquire affordance by retrieving objects that share visual or semantic similarities from the affordance memory. The next step is to map the contact points of the retrieved objects to the new object. While establishing this correspondence may present formidable challenges at first glance, recent research finds it naturally arises from pre-trained diffusion models, enabling affordance mapping even across disparate object categories. Through the \method framework, robots may generalize to manipulate out-of-category objects in a zero-shot manner without any manual annotation, additional training, part segmentation, pre-coded knowledge, or viewpoint restrictions. Quantitatively, \method significantly enhances the accuracy of visual affordance retrieval by a large margin of \textbf{31.6\%} compared to state-of-the-art (SOTA) end-to-end affordance models. We also conduct real-world experiments of cross-category object-grasping tasks. \method achieved a success rate of \textbf{85.7\%}, proving its capacity for real-world tasks.
\end{abstract}

%-------------------------------------------------------------------------
%%%%%%%%% BODY TEXT
\vspace{-0.2cm}
\section{Introduction}
%-------------------------------------------------------------------------
Imagine a future where robots assist humans to proficiently accomplish a broad spectrum of daily tasks. The fundamental challenge lies in empowering robots to find interaction strategies with both familiar and novel objects. We humans instinctively have such abilities by generalizing the affordance~\cite{gibson1978ecological} to unseen objects through semantic mapping~\cite{creem2005neural}. For example, we may figure out how to grab a badminton racket by recalling the experience of wielding a tennis racket, or how to open a microwave oven that partially resembles a cabinet. 

However, this ability is not innate to robots. A key challenge is to obtain such interaction experience and extract generalizable information for robot manipulation. Fortunately, there is a wealth of egocentric human-object interaction videos~\cite{damen2020epic,grauman2022ego4d,damen2018scaling} available on the Internet. These videos provides valuable insights into complex interactions, as well as the temporal and motion contexts of objects. 
Previous works have explored a variety of methods for learning object affordances from videos, such as extracting feature embeddings from videos~\cite{demo2vec2018cvpr}, or automatically collecting pseudo-ground-truth labels for end-to-end training~\cite{bahl2023affordances,bahl2022human,liu2022joint}. While existing methods can predict affordance for familiar objects, they struggle to generalize to unseen objects. 

We aim to effectively and efficiently generalize affordance beyond object categories. To this end, we propose a general framework, \method, that can recall object interaction experience from human videos and transfer it to novel objects. First, we extract the interaction experiences of objects from human videos and store them in an affordance memory. Second, in the face of a novel object  (\textit{e.g.} a screwdriver), we retrieve objects (\textit{e.g.} a knife) similar to the target object from the memory based on visual and semantic similarity. The most intriguing aspect of our method is the third step, where we employ the emergent semantic correspondence ability from the diffusion models~\cite{tang2023dift, hedlin2023unsupervised, luo2023dhf, zhang2023tale} to map the retrieved contact points to the novel object. We find this procedure powerful enough to transfer affordance beyond multiple object categories. Finally, we use the obtained affordance points to select from the grasping position prior provided by AnyGrasp\cite{fang2023anygrasp} and deploy on a real robot to complete the manipulation task.
%exp summary
We conduct a comprehensive evaluation of \method's generalization ability under different settings. We evaluate the zero-shot generalization ability with other end-to-end methods, and the affordance prediction success rate of Robo-ABC significantly increases by 31.6\%, demonstrating the strong generalization capability to novel objects and categories. We also demonstrate \method's potential to generalize the affordance of one source object to objects that span a large category gap. Lastly, after deploying the \method on a real robot using AnyGrasp~\cite{fang2023anygrasp}, we show that our method can provide accurate affordance guidance for grasping in open-world, novel view, and cross-category settings. \method reaches a prominent success rate of 85.7\% over 7 object categories.
% ...\todo{different categories task}

Our contributions can be summarized as follows:

1) We propose a novel framework, \method, to extract object interaction experience from human videos and transfer it to novel objects with no need for annotation, additional training, or pre-coded knowledge of any kind.

2) We demonstrate the effectiveness of our method in zero-shot affordance generalization and cross-category generalization settings, achieving a significant improvement of 31.6\% in success rate over previous end-to-end methods.

3) We believe that the pipeline of \method is versatile and naturally enjoys the benefit of the increasing ability of foundation models, both in retrieving similar objects and capturing semantic correspondence. We hope that our work will inspire future research in this direction.

\section{Related Works}
\begin{figure*}[t]
    \centering
    \vspace{-0.9cm}
    \includegraphics[width=1.0\textwidth]{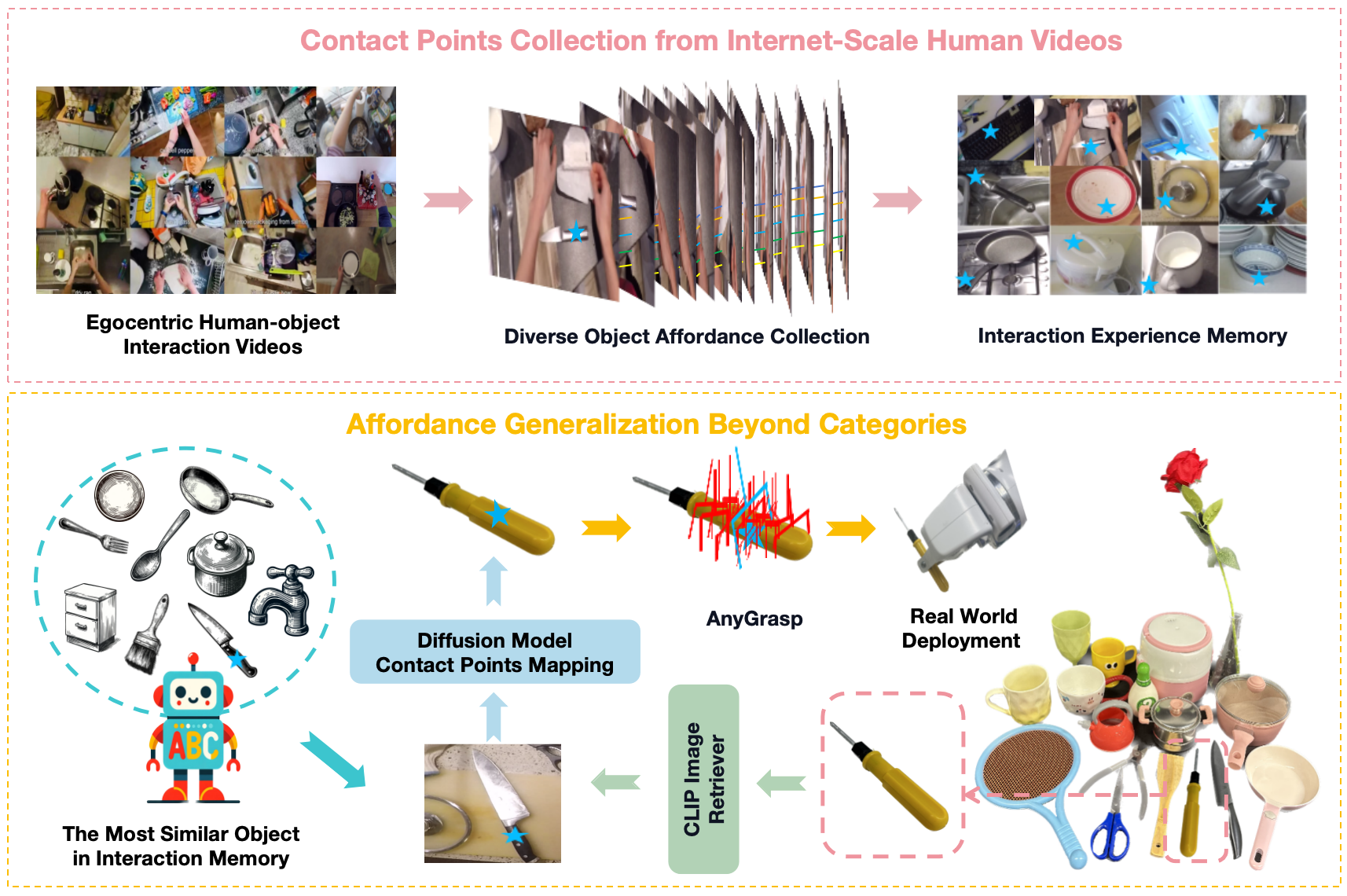}
    \vspace{-0.6cm}
    \caption{\textbf{Our pipeline.} The top part is the process of extracting knowledge about object affordance from human-object videos. Subsequently, we store these information as interaction memory to serve as the robot's interaction experience. When facing new objects, we retrieve the most similar object from the interaction memory based on visual and semantic similarity. After obtaining the contact point information, we leverage the powerful semantic correspondence capability in the diffusion model to achieve cross-object and out-of-category affordance generalization. Finally, we select the grasp pose from  all the possible poses which are generated by AnyGrasp~\cite{fang2023anygrasp} to deploy on real robots.\ (\textcolor{brightblue}{\Pisymbol{pzd}{72}}  represents the positions for interacting with the object,  \ \usebox{\mycustomr}\ represents all possible grasp poses generated by AnyGrasp, \ \usebox{\mycustomb}\  represents the grasp pose selected by \textcolor{brightblue}{\Pisymbol{pzd}{72}} )
    % mainly consists of five stages. First, we extract interaction locations from human video datasets, treating such affordance information as the robot's memory of the affordances of seen objects. When facing new objects, for seen categories, we retrieve the most similar object within the category based on visual and semantic similarity. For unseen categories, we search among all stored categories. After obtaining the contact point information, we leverage the powerful semantic correspondence capability in the diffusion model to achieve cross-object and cross-category correspondence of affordances. Finally, we use AnyGrasp~\cite{fang2023anygrasp} to perform operational tasks on real robots.
    }
    \vspace{-5mm}
    \label{overview}
\end{figure*}
%-------------------------------------------------------------------------
\subsection{Learning visual affordance from human videos}
\label{sec:visual_affordance_from_human_videos}
%----------------------------------------------------------------------
Visual affordance learning aims to infer where and how to interact with diverse objects from visual inputs, bridging the computer vision and robotics fields. RGB image-based affordance oklearning~\cite{luo2022learning,hou2021affordance,zhu2014reasoning,hassan2016attribute,ye2023affordance,myers2015affordance} methods focus on inferring affordances from images depicting human-object interactions.  Another series of works focus on predicting affordance from 3D point cloud inputs~\cite{mo2021where2act,mo2022o2o,wu2021vat,wang2022adaafford,cheng2023learning,ning2023where2explore,geng2023rlafford,zhao2022dualafford}, specifically targeting articulated objects manipulation. While taking the RGB image as sensory input, our work focuses on extracting affordance from egocentric human videos~\cite{damen2018scaling,damen2022rescaling,damen2020epic,grauman2022ego4d}, which can capture the temporal context and motion information of human-object interaction. These allow us to better understand complex actions and generalize to new objects and scenarios.
Based on the diverse reservoir of human videos, previous works have investigated to learn from them the visual representation~\cite{nair2022r3m,bahl2023affordances}, grasp prior~\cite{kannan2023deft,mendonca23swim,bahl2022human,mandikal2022dexvip}, and dexterous grasping skills~\cite{mandikal2021learning,wu2023learning}. 

The most relevant studies to our goal are ~\cite{goyal2022human,nagarajan2019grounded,liu2022joint,bahl2023affordances,li2022discovering}. These works are dedicated to identifying the contact region with objects from human videos. However, these end-to-end approaches to affordance prediction are subjected to in-domain object instances and viewpoints. Compared to previous works, \method extracts a small-scale memory of object interaction experiences from human videos. It allows the robot to face completely new scenes with transferable manipulation knowledge. What distinguishes our method from the previous ones is our focus on capturing the semantic correspondence beyond the seen object categories to help guide robots more accurately in complex zero-shot manipulation tasks.
%-------------------------------------------------------------------------
\subsection{Semantic correspondence for robotics}
%----------------------------------------------------------------------
In the robotics field, previous works ~\cite{wang2023d,xue2023useek,jiang2023doduo,di2024effectiveness} have explored capturing semantic correspondences for robot manipulation. However, these works are somewhat limited to generalization within different instances of the same category, additional training or rely on user-provided goal images to perform the transfer. In this work, our goal is to achieve zero-shot generalization across object categories. Recently, foundational models such as DINO-VIT~\cite{amir2021deep} and diffusion models~\cite{hedlin2023unsupervised,zhang2023tale,tang2023emergent} have demonstrated remarkable capabilities in finding semantic correspondences across objects. Specifically, features extracted from diffusion models are more versatile in mapping similar points across categories. As foundation models contain knowledge valuable for robotics tasks~\cite{gao2023can, ye2023foundations}, we explore leveraging the semantic correspondence knowledge embedded within these models, eliminating the need for additional training or part segmentation. Compared to previous methods, our approach is off-the-shelf and achieves significant improvements. This provides better visual guidance information for robot manipulation tasks, allowing robots to flexibly understand and infer the affordances of different categories of objects in the open world.
\begin{figure*}[t]
    \centering
    \vspace{-0.5cm}
    \includegraphics[width=1.0\textwidth]{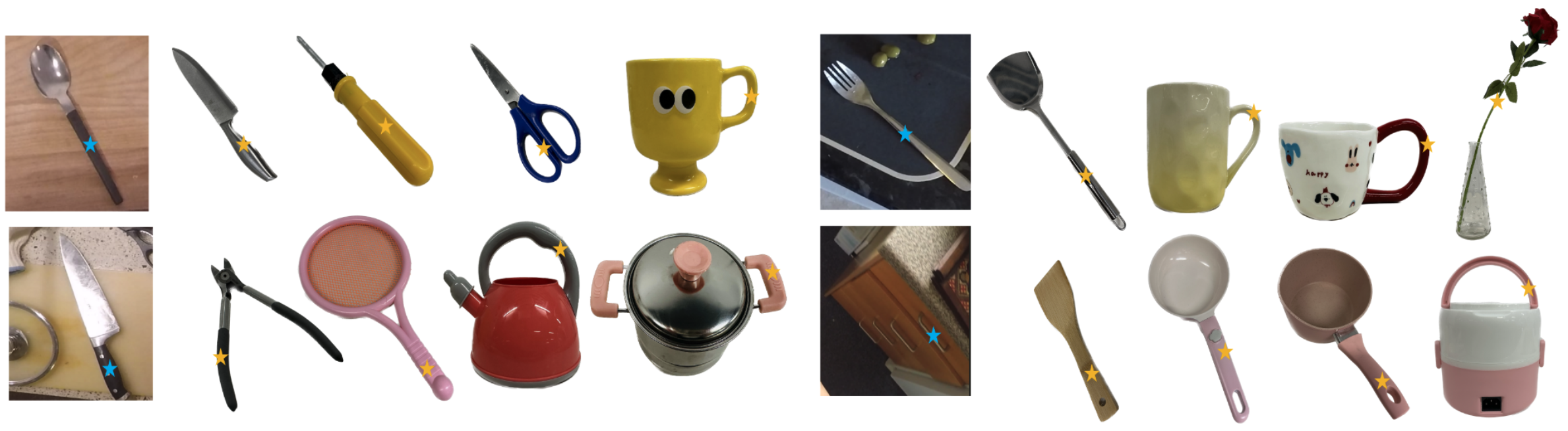}
    \vspace{-0.7cm}
    \caption{\textbf{Affordance generalization beyond categories visualization results.} In each group of figures from left to right, the span of object categories gradually increases. \textcolor{brightblue}{\Pisymbol{pzd}{72}} represents the contact points extracted from human videos, while \textcolor{orange}{\Pisymbol{pzd}{72}} represents the inferred points found by Robo-ABC across object categories. 
    }
    \vspace{-5mm}
    \label{fig:all_types}
\end{figure*}
%-------------------------------------------------------------------------
%{Cross-Category Generalizable Robotic Manipulation}
%----------------------------------------------------------------------
\subsection{Generalizable robot manipulation}
In the development of general-purpose robots, having generalizable manipulation capabilities is crucial, especially when applying these abilities across object categories. Facing this challenge, some works\cite{geng2023gapartnet,geng2023partmanip} focus on using point clouds as inputs, recognizing and manipulating actionable parts to achieve cross-category object manipulation. However, these methods typically require a large amount of annotated data or rely on effective part segmentation of the object. 
Recent works on general agents \cite{brohan2023can,reed2022generalist,xue2023arraybot} and dexterous grasping \cite{li2023gendexgrasp,xu2023unidexgrasp,wan2023unidexgrasp++,li2023vihope} continue to explore new possibilities. Another line of research \cite{rashid2023language, wang2023d} utilizes foundation models and NeRF for generalizable manipulation. However, the manipulation skills of these agents are confined to a set of known instances, and their ability to generalize falls short when encountering novel object instances. By contrast, our key motivation is to harness foundation models to extract semantic information and achieve affordance generalization beyond categories.
%------------------------------------------------------------------------
\section{Method}
In this study, we aim to obtain contact points for robotic manipulation, with a focus on generalizing to unseen objects and categories. The structure of this section unfolds as follows: Section \ref{sec:affordance_collection} elucidates the process of extracting affordance knowledge from human videos, Section \ref{sec:retrieve_reference_frame} discusses the retrieval of similar objects from the extracted interaction experience memory for new objects, Section \ref{sec:semantic_correspondence} describes the process of utilizing the capabilities of the diffusion model to generalize the affordance of objects across object instances and categories, while Section \ref{sec:real_world} describes the process of applying the obtained affordance guidance to downstream robotic manipulation tasks.
\begin{figure*}[t]
    \centering
    \vspace{-0.5cm}
    \includegraphics[width=1.0\textwidth]{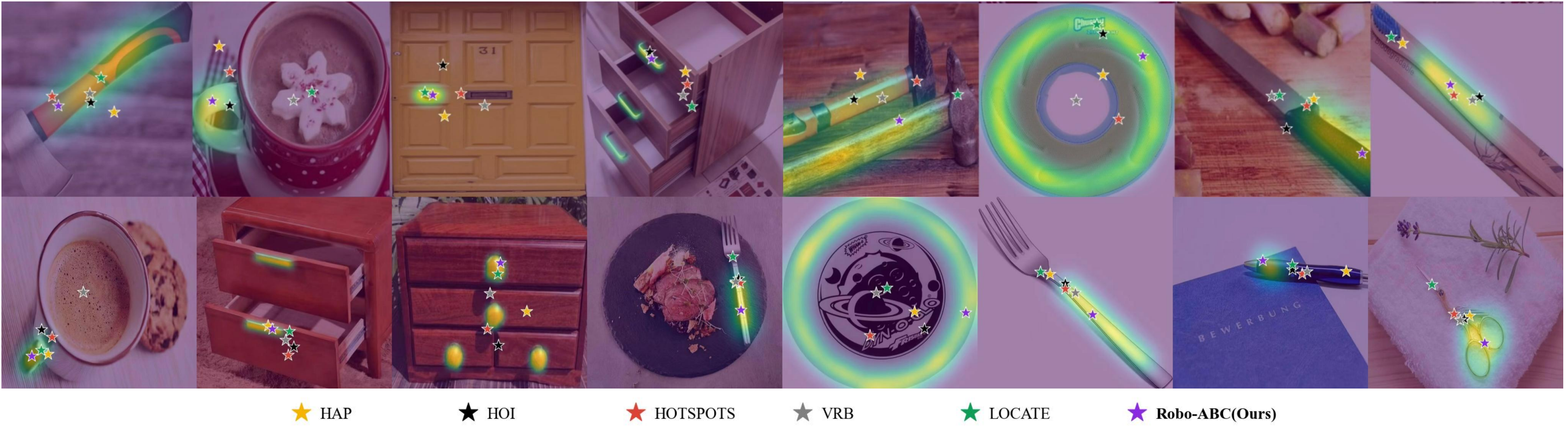}
    \vspace{-0.5cm}
    \caption{\textbf{Visualization of the affordance results. } The highlighted areas are the ground truth masks, while \textcolor{orange}{\Pisymbol{pzd}{72}} \textcolor{black}{\Pisymbol{pzd}{72}} \textcolor{brightred}{\Pisymbol{pzd}{72}} \textcolor{gray}{\Pisymbol{pzd}{72}} \textcolor{brightgreen}{\Pisymbol{pzd}{72}} \textcolor{brightpurple}{\Pisymbol{pzd}{72}} indicate the predicted contact points of different methods.
    }
    \vspace{-1mm}
    \label{fig:category_success}
\end{figure*}
\subsection{Affordance collection from human videos}
\label{sec:affordance_collection}
\subsubsection{Affordance representation}
We define the affordance of objects as the contact points between the human hand and the object. Humans naturally contact a variety of objects at specific points during the interaction. For example, when opening a door, the hand contacts the door handle, and such action bears fruitful information on affordance. We aim to pinpoint \textbf{\textit{when}} and \textbf{\textit{where}} the contact takes place given a dataset of videos. We utilize an off-the-shelf hand-object interaction detector~\cite{shan2020understanding} to obtain the grasp state information during human-object interaction in each segment of the interaction videos. 

Consider a video consisting of multiple frames of a person cutting vegetables: $V = \{F_1, ..., F_N\}$. We first employ a hand-object detector to determine whether the hand and the object are contacted in each frame, as well as to obtain the pixel-level bounding box (bbox) of the hand $B_h$ and the object $B_o$. Upon identifying the first frame $F_j$ where the hand grasps the knife, we use skin segmentation ~\cite{saxen2014color} to precisely locate the intersection area within the hand bbox $B_h$ and objects bbox $B_o$. We then randomly sample from this area to obtain the contact points $P = \{p_1, p_2,...,p_m\}$, where $m$ is the sample number of contact points in a frame.

\subsubsection{Contact points mapping}
\label{sec:contact_point_mapping}

When collecting affordance information, we want the image to be clear, preferably without humans occluding the object.
However, we need the frame of hand-object contact to deduce the contact point. %we need to consider several issues. Firstly, there exists a morphological difference, or a domain gap, between humans and robots. Secondly, for subsequent similar object retrieval, we require images of the complete object, unobscured. Thirdly, when seeking semantic correspondence between different objects, we need clear images. Otherwise, it could affect the accuracy of the correspondence. Therefore, these considerations necessitate careful handling of affordance data collecting.
To address this dilemma, we aim to map these contact points $P$ back to a frame $F_c$ when the object is not obscured. 
This can be achieved through calculating the homography matrix $\mathcal{H}_t$ between two consecutive frames to map the contact points $P$ across different frames.

There exist several criteria for the selection of $F_c$: Since these frames are extracted from a video, preventing motion blur in these frames are cruial. It is also preferable to retrieve the frames near the contact frame $F_j$, so we can reduce the mapping error. With these considerations in mind, we set a window $W$ around the contact frame $F_j$ to select the frame $F_c$ where the object's view is intact. For motion blur detection, we utilize the Laplacian operator to compute the clearest frame within the window $W$. We then calculate the homography matrix to map the contact point to the unobstructed frame.
Lastly, we use the bounding box of the object $B_o$ output by the hand-object detector ~\cite{shan2020understanding} to get rid of the irrelevant surroundings of the object. We collect these cropped object images $I_o$ and contact points $P$ to store in the affordance memory, serving as the robot's knowledge bank of interaction experiences.

\subsection{The most similar object retrieval from memory}
\label{sec:retrieve_reference_frame}
When facing a new object, we need to retrieve similar objects from the collected memory. After capturing an image $I_t$ with the camera, we use langsam \cite{luca2023lang} to crop the object we want to interact with from the image. An image encoder $\Phi$ is used to map the cropped object image to a feature vector $z_{\text{crop}}$. We also map each object image in the memory to a feature vector $z_{\text{mem}}$ using the same encoder $\Phi$. We use the cosine similarity of these two feature vectors as the proximity metric for retrieval.

We divide these objects into two types: The first type includes objects within the same category that have been previously encountered in the collected interaction memory, and the second type includes objects that are unfamiliar or have never been seen before. For the seen objects, we retrieve the most similar objects in the same category. For completely new object categories, we retrieve the most similar object among all objects in the memory, regardless of the category. In practice, we find that the CLIP \cite{Radford2021LearningTV} encoder suitably meets our needs, and the effects of other encoders have been tested and demonstrated in Section \ref{sec:ablation}. 

\subsection{Semantic correspondence mapping for affordance generalization}
\label{sec:semantic_correspondence}

After retrieving the most similar object in the memory and the contact points of it, we utilize semantic correspondence mapping to transfer such knowledge to the current scene and object. Semantic correspondence maps points in the source image to the target image. In this work, we utilize the emergent semantic correspondence ability from the diffusion model to map the retrieved contact points to the new object, thereby guiding the robotic manipulation task in unfamiliar environments. 

More specifically, given a source image $I_s$, a target image $I_t$, and a source point $p_s$, we aim to find the corresponding point $p_t$ in the target image. We follow the steps described in \cite{tang2023dift} to extract the diffusion features (DIFT) of the source image $I_s$ and the target image $I_t$. The diffusion features are generated by first adding noise to the goal image, then denoising it through the diffusion process, while extracting the intermediate hidden features from the U-Net simultaneously. We refer the readers to the original paper for more details.

Since the diffusion features correspond to each pixel in the goal image, we can find the pixel with the highest similarity to the source point $p_s$ in the diffusion features of the target image $I_t$. %This simple but effective method allows us to map the contact points of the retrieved object to the current object with high fidelity and can even break the barrier caused by categories, viewpoints, and background appearance. 
Specifically, we find that diffusion features are relatively prone to orientation mismatching between the source and the target image, so we deploy 8 rotation and flipping transformations to the source image, and select the result with the highest similarity among all transformed images. While we also tried other semantic correspondence approaches, as discussed in Section \ref{sec:ablation}, we found that the diffusion model is the most effective for our task.
\subsection{Deployment in the real world}
\label{sec:real_world}
After obtaining the contact points of the current object, we deploy on a real robot to complete the manipulation tasks. We utilize the AnyGrasp~\cite{fang2023anygrasp} to provide the grasping prior for the robot. AnyGrasp ~\cite{fang2023anygrasp} can take the point cloud of the scene and provide a set of 7-DOF grasp candidates for grasping. We utilize the contact points obtained from the previous step to %filter out the grasp candidates that are not suitable for the current object, 
select the nearest grasp candidate as the final grasp pose, and use it to guide the robot to complete the manipulation task. We will discuss the details of the deployment in Appendix \ref{app:robot_hardware}. %\kz{add ref and details..}.

\section{Experiments}

In this section, we present a comprehensive evaluation of \method, and specifically address the following questions: 1) How does the zero-shot affordance generalization ability of \method compare to existing end-to-end approaches for both familiar and novel object categories? 2) To what extent can \method extrapolate affordances from a limited set of known categories to a broader spectrum of objects? 3) Upon implementation on the robot, how accurate is the semantic information provided by \method for affordance-guided grasping, particularly in scenarios involving varying viewpoints and novel object categories?

We will elaborate on these questions in the following sections. By conducting these experiments, we hope to provide a comprehensive evaluation of our method's affordance generalization ability and shed light on potential areas for further research and improvement.
%----------------------------------------------------------------------
\begin{figure*}[t]
    \centering
    \vspace{-0.5cm}
    \includegraphics[width=1.0\textwidth]{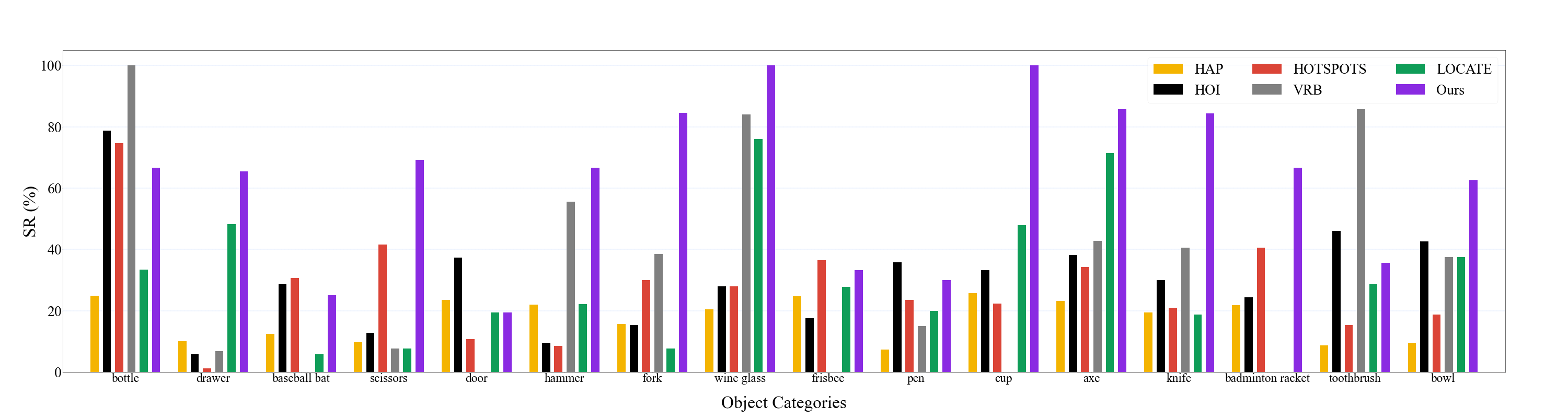}
    \vspace{-0.5cm}
    \caption{\textbf{Success rate by category. } We demonstrate the performance of Robo-ABC and other baselines across various object categories within the entire evaluation dataset. As can be seen, in the vast majority of cases, Robo-ABC exhibits superior zero-shot generalization capabilities.
    }
    \vspace{-1mm}
    \label{fig:category_success}
\end{figure*}

\subsection{Zero-shot affordance generalization}
\vspace{-0.2cm}
\subsubsection{Evaluation Dataset}
\label{sec:evaluation_dataset}
In this experiment, we select all the objects that are feasible for robotic manipulation from the “hold” action category of the AGD20K~\cite{luo2022learning} dataset, ruling out objects that are too large for the robot to grasp such as beds and chairs. Considering the prevalence of doors and drawers in everyday life, we supplemented the evaluation set with these two additional categories following the same labeling procedure. All object categories within the evaluation data set are shown in Table \ref{tab:seen_unseen_objects}. We use the terms \textbf{seen categories} and \textbf{unseen categories} here to refer to whether objects of the same category are present in the affordance memory. %\kz{Compared to other baselines, our method only extracts small-scale affordance information of objects. The amount of data used by other baselines is much larger than ours, and the types of objects in our memory are fully included in the datasets used by other baselines. (need to be more specific)}
\begin{table}[ht]
\centering
\caption{For the list of seen and unseen objects in the evaluation dataset of the affordance memory}
\label{tab:objects}
\begin{tabularx}{.45\textwidth}{@{}>{\centering\arraybackslash}m{1.5cm}X@{}}
\toprule
\textbf{Seen}   &  bottle, bowl, drawer, cup, door, 
    fork\\  & knife, scissors, wine glass \\
\midrule
\textbf{Unseen} & axe, badminton racket, baseball bat, \\
                & frisbee, hammer, pen, toothbrush \\
\bottomrule
\end{tabularx}
\label{tab:seen_unseen_objects}
\end{table}
\vspace{-0.5cm}
\subsubsection{Evaluation Metrics}
\label{sec:evaluation_metrics}
In choosing the evaluation metrics, we claim that for the purpose of real-world object manipulation, the model's output should be contact points, instead of areas resembling a probability distribution. Therefore, we compare the accuracy of the predicted contact points across different methods. As such, we selected three metrics for our evaluation. Detailed explanations are as follows:

\noindent\textbf{Success Rate (SR): } The success rate is calculated as the proportion of successful points (those falling within the ground truth (GT) masks) to all points. Since the value of the ground truth mask ranges from 0 to 255, we set a threshold of 122 to determine whether the output is feasible while leaving the SR-threshold curve for the appendix. The success rate measures the accuracy of the output directly, making it the primary metric for affordance generalization. 
% The formula of SR is as follows:
% \[
% SR = \frac{\text{Number of successful points}}{\text{Total number of output points}}
% \]

\noindent\textbf{Normalized Scanpath Saliency (NSS):}
Normalized Scanpath Saliency is a straightforward method to measure the correlation between saliency maps and fixed points, calculated by averaging the normalized saliency at points of the ground truth. Since we are dealing with predicted points and ground truth maps, we have modified the formula to compute the average normalized value of the ground truth map at the output points. For ground truth map $\mathcal{M}$ and output points $P$, the formula of NSS is as follows:

\begin{equation}
NSS = \frac{1}{|P|} \sum_{p \in P} \frac{\mathcal{M}(p)}{\max_{q \in \mathcal{M}} \mathcal{M}(q)} \in [0, 1]
\end{equation}

NSS considers not only the accuracy of the output points but also their saliency. The higher the NSS, the more accurate the output points are to the center of the ground truth map.

\noindent\textbf{Distance to Mask (DTM):}
We introduce a novel metric to compute the shortest distance between the ground truth mask region and the predicted affordance position. Using the same threshold as in success rate, we can obtain the contour $\mathcal{C}_\mathcal{R}$ of the ground truth mask $\mathcal{M}$. If the output point $P$ is outside the mask area, we calculate the shortest distance from $P$ to the contour $\mathcal{C}_\mathcal{R}$. If the output point $P$ is inside the contour $\mathcal{C}_\mathcal{R}$, the distance to the mask is 0. DTM is then normalized by the length of the image's diagonal. 
\vspace{-0.4cm}
\subsubsection{Baselines and results} 

For the zero-shot affordance generalization experiments, we compare \method with a series of previous end-to-end approaches, namely VRB~\cite{bahl2023affordances}, HOI~\cite{liu2022joint}, HOTSPOTS~\cite{nagarajan2019grounded}, and HAP~\cite{goyal2022human}. Additionally, we also evaluate LOCATE~\cite{luo2022learning}, the work that proposed the AGD20K dataset on our benchmark. A brief introduction to these methods can be found in Appendix~\ref{sec:affordance_model}.

As we stated earlier, what we want to compare is the accuracy of the contact points rather than probabilistic distributions. Thus, we have constrained all baselines to predict points for a fair comparison. For models that output heatmaps, we select the points with the top 5 probability. 

As shown in Table \ref{tab:main_table}, \method achieves high success rate of 60.7\%, which is 31.6\% higher than the second-best method, LOCATE. This demonstrates the effectiveness of our method in generalizing affordance to unseen objects. We also observe that the NSS and DTM of \method are significantly better than other methods, indicating that the results are closer to the center of the ground truth mask. 
\vspace{-0.2cm}
\begin{table}[h]
\renewcommand{\arraystretch}{1.2}
\setlength\tabcolsep{10pt}
\centering
\caption{\textbf{Main results on affordance prediction.} \method surpasses all baselines by a large margin on all three metrics}
{\begin{tabular}{lcccccc}
\toprule
\multicolumn{1}{c}{\bfseries{Methods}} & \multicolumn{1}{c}{NSS} & \multicolumn{1}{c}{SR} & \multicolumn{1}{c}{DTM} \\
\hline
\multicolumn{1}{c}{HAP\cite{goyal2022human}} & \multicolumn{1}{c}{0.231} & \multicolumn{1}{c}{22.4} & \multicolumn{1}{c}{0.121} \\
\multicolumn{1}{c}{HOI\cite{liu2022joint}} & \multicolumn{1}{c}{0.239} & \multicolumn{1}{c}{26.1} & \multicolumn{1}{c}{0.112} \\
\multicolumn{1}{c}{HOTSPOTS\cite{nagarajan2019grounded}} & \multicolumn{1}{c}{0.236} & \multicolumn{1}{c}{23.6} & \multicolumn{1}{c}{0.118} \\
\multicolumn{1}{c}{LOCATE\cite{luo2022learning}} & \multicolumn{1}{c}{0.283} & \multicolumn{1}{c}{29.1} & \multicolumn{1}{c}{0.107} \\
\multicolumn{1}{c}{VRB\cite{bahl2023affordances}} & \multicolumn{1}{c}{0.242} & \multicolumn{1}{c}{26.9} & \multicolumn{1}{c}{0.103} \\
\multicolumn{1}{c}{\bfseries{Robo-ABC (Ours)}} & \multicolumn{1}{c}{\bfseries{0.516}} & \multicolumn{1}{c}{\bfseries{60.7}} & \multicolumn{1}{c}{\bfseries{0.045}} \\
\bottomrule
\end{tabular}}
\label{tab:main_table}
\end{table}
\vspace{-0.5cm}
\subsection{Cross-Category Affordance generalization}

In this experiment, we aim to showcase our method's ability to generalize the affordance of a small group of seen objects to various objects beyond its category. To this end, we fix a category of source images and provide the contact points derived from human videos. For each object of the other category, we use the same semantic correspondence setting of \method, then obtain the target affordance, as shown in Figure \ref{fig:all_types}. The similarities between the source category to the target are gradually decreasing from left to right, demonstrating the increasing challenge of affordance generalization across objects. The correspondence model of \method can generalize even under the most challenging circumstances, like mapping the handle of a fork to the stem of a flower, which infers the great potential of our method in generalizing affordance across object categories.

\begin{table*}
\renewcommand{\arraystretch}{1.2}
\setlength\tabcolsep{6.2pt}
  \centering
  \caption{The performance of different retrieval methods under various correspondence models.}
  \begin{tabular}{lcccccccccccc}
    \toprule
    \multirow{2.5}{*}{\bfseries{Correspondence}}&\multicolumn{12}{c}{\bfseries{Retriever Encoder}}\\
    \hhline{~------------}
    \multirow{2.65}{*}{\bfseries{Methods}}&\multicolumn{3}{c}{\yellow CLIP-B32} & \multicolumn{3}{c}{\purple CLIP-B32+LPIPS} &\multicolumn{3}{c}{\blue CLIP-B16} & \multicolumn{3}{c}{\pink ResNet50}\\
     \hhline{~------------}
    &\yellow DTM&\yellow NSS&\yellow SR&\purple DTM&\purple NSS
 &\purple SR&\blue DTM&\blue NSS&\blue SR&\pink DTM&\pink NSS&\pink SR\\
    \hline
    \hline
    \multicolumn{1}{l|}{DIFT~\cite{tang2023dift}} &\yellow \textbf{0.045}&\yellow 0.516&\yellow \textbf{60.7}&\purple \textbf{0.067}&\purple \textbf{0.438}&\purple 50.2&\blue \textbf{0.052}&\blue \textbf{0.522}&\blue \textbf{60.0}&\pink \textbf{0.058}&\pink 
    \textbf{0.495}&\pink \textbf{56.7}\\
    \multicolumn{1}{l|}{SD-DINO~\cite{zhang2023tale}} &\yellow 0.052&\yellow \textbf{0.524}&\yellow 58.2&\purple 0.080&\purple 0.387&\purple \textbf{54.2}&\blue 0.077 &\blue 0.418 &\blue 55.6 &\pink 0.138&\pink 0.360&\pink 50.5\\
    \multicolumn{1}{l|}{DINO-VIT~\cite{carion2020end}}&\yellow 0.158&\yellow 0.247&\yellow 36.4&\purple 0.153&\purple 0.255&\purple 37.1&\blue 0.138&\blue 0.296&\blue 41.8&\pink 0.160&\pink 0.209&\pink 28.7\\
    \multicolumn{1}{l|}{LDM-SC~\cite{hedlin2023unsupervised}}&\yellow 0.087&\yellow 0.390&\yellow 52.0&\purple 0.106&\purple 0.385&\purple 50.9&\blue 0.096&\blue 0.408&\blue 53.5&\pink 0.141&\pink 0.304&\pink 40.7\\
    \bottomrule
  \end{tabular}
  \label{tab:retrieve_correspondence}
\end{table*}

\subsection{Real-World robot experiment}
%Lastly, we plan to deploy our method and the two best-performing baselines from Experiment 1 in real-world scenarios. This will allow us to evaluate the performance and generalization ability in practical situations. It will also offer valuable feedback for further refinements and improvements of our method.
Lastly, we deploy \method along with the VRB baseline in real-world scenarios. Our method is adept at generating semantically-informed contact points. Combined with the end-to-end grasping backbone AnyGrasp, we can conduct experiments of grasping with a variety of categories.

 AnyGrasp~\cite{fang2023anygrasp} is trained on a large number of real-world grasping scenarios, enabling it to generate robust and reliable grasp proposals. It takes as input a point cloud from the depth camera and outputs a set of 7-DOF grasp poses. Each detected grasp pose $G$ is represented as $g$ = [$\mathbf{R}\in \mathbb{R}^{3 \times 3}$, $\mathbf{t}\in \mathbb{R}^{3 \times 1}$, $w\in \mathbb{R}$], where $\mathbf{R},\mathbf{t},w$ signify the rotation, translation, and width of the gripper, respectively.
 
 Our robotic setup consists of a Franka arm equipped with a parallel jaw gripper, and a RGBD camera mounted to provide a monocular point cloud of the scene. 
 
 Initially, we compute the contact point for the given object and ascertain its three-dimensional spatial coordinates $p^* = (x, y, z)$. The scene's point cloud is then fed into AnyGrasp to generate a set of grasp candidates $G$:\{$g_1, g_2, ..., g_N$\}. Among these, the pose $g^*$, exhibiting the minimal translational distance from point $p^*$, is selected as the execution pose for the robot's end-effector.

\begin{equation}
g^* = \mathop{\arg\min}\limits_{g \in G} ||\mathbf{t}(g) - p^*|| 
\end{equation}

Our experiments are conducted over seven object categories. \method achieves a success rate of 85.7\%, compared to the baseline of 68.6\%. Please refer to Appendix \ref{app:robot_hardware} and \ref{app:robot_results} for more details.
%----------------------------------------------------
\vspace{-3mm}
\section{Ablation Study}
\label{sec:ablation}
In this section, we examine several implementation choices of \method, validating their influences by controlling variables. The selected choices include the number of seen object categories for the afforance memory, the retriever encoder, the number of retrieved images, and the selection of semantic correspondence models. 
\vspace{-0.3cm}
\subsection{Retriever \& Semantic correspondence model}
\label{sec:retriever_semantic_correspondence_model}

For the retriever, we compare the performance of four different encoders: CLIP-B32, CLIP-B32+LPIPS, CLIP-B16, and ResNet50. The three encoders with the prefix CLIP are all based on the CLIP~\cite{radford2021learning} model with different parameter sizes, and the last one is based on the ResNet50 model. Specifically, the CLIP-B32+LPIPS retriever uses the CLIP-B32 encoder with the negative LPIPS~\cite{zhang2018perceptual} loss added to the CLIP similarity, aimming to fetch the most visually and semantically similar images from the memory.

For the semantic correspondence model, we compare the performance of four different models: DIFT~\cite{tang2023dift}, SD-DINO~\cite{zhang2023tale}, LDM-SC~\cite{hedlin2023unsupervised}, and DINO-VIT~\cite{carion2020end}. While the first three correspondence methods are based on the diffusion models, the last one is based on the VIT. We refer the readers to Appendix \ref{sec:semantic_correspondence_app} for more details.

The results are shown in Table \ref{tab:retrieve_correspondence} with all the metrics, and we can see that the CLIP-B32 encoder achieves the best performance, and the CLIP-B32+LPIPS encoder is slightly weaker. This indicates that the CLIP-B32 encoder alone can already provide decent results. 
% In addition, we can see that the CLIP-B16 encoder is slightly worse than the CLIP-B32 encoder, and the ResNet50 encoder is the poorest. 
% This shows that the CLIP encoder is more suitable for our task than the ResNet50 encoder. 
For semantic correspondence methods, the DIFT\cite{tang2023dift} model achieves the best performance, and the DINO-VIT\cite{carion2020end} model is the weakest, While the other two models are slightly worse than the DIFT~\cite{tang2023dift} model. % This result indicates that the diffusion model is more suitable for our task than the DINO feature.
\subsection{Memory size of seen categories}

Based on the results of the previous ablation study, we select the combination of CLIP-B32 retriever with DIFT~\cite{tang2023dift} for correspondence matching. In this experiment, we fix the encoder of the retriever, aiming to validate the impact of the size of the “seen categories” on different semantic correspondence models. The results are shown in Table \ref{tab:memory_category}.
\begin{table}[h]
\renewcommand{\arraystretch}{1.2}
\setlength\tabcolsep{10pt}
\centering
\caption{The impact of categories seen in memory on the semantic correspondence performance of different models.}
  {\begin{tabular}{lccc}
  \toprule
  \multirow{1.1}{*}{\bfseries{Correspondence}}& \multicolumn{3}{c}{\bfseries{Memory Categories}}\\
   \hhline{~---}  
  \multirow{1.25}{*}{\bfseries{Methods}}& 18 & 36 & 51 \\ 
  \hline
  \hline
   \multicolumn{1}{l|}{DIFT~\cite{tang2023dift}}& 56.4& 56.7& \textbf{60.7}\\
   \multicolumn{1}{l|}{SD-DINO~\cite{zhang2023tale}}& 53.1& 53.1& 58.2\\
   \multicolumn{1}{l|}{DINO-VIT~\cite{carion2020end}}& 39.3& 37.5& 36.4\\
  \bottomrule
  \end{tabular}}
  \vspace{-0.3cm}
  \label{tab:memory_category}
\end{table}

The hypothesis is that having more variety in the seen categories may improve the model's ability to generalize and correctly match contact points. However, it is also possible that there is an optimal number of categories, beyond which model performance may not improve or even degrade because the retriever is disturbed by a large number of categories and unable to retrieve the most relevant images.

By varying the number of seen categories while keeping other variables constant, we can determine the optimal number of object categories for the model to learn from, which is crucial information for improving the system's overall performance. From the results in Table ~\ref{tab:memory_category}, we can see that the performance of the more capable models (DIFT and SD-DINO) increases as the number of seen categories increases. This result indicates that the more categories the model sees, the better the performance. However, for the less capable model of DINO-VIT, the performance peaks at 36 categories and then decreases. 
% \vspace{-2mm}
\subsection{Number of retrieved images}
In this experiment, we aim to validate the impact of the number of retrieved top-k images from the retriever on the performance of the semantic correspondence model. The image with the highest DIFT similarity is selected for the predicted contact point. The results are shown in Table \ref{tab:top_k}. We can see that the performance of the three models increases as the number of retrieved images increases, the performance may further improve shall we retrieve more images, but this is at the cost of longer inference time.
\vspace{-0.3cm}
\begin{table}[h]
\renewcommand{\arraystretch}{1.2}
\centering
\caption{The performance of different models in terms of the number of retrieved images.}
\label{tab:top_k}
\setlength\tabcolsep{10pt}
  {\begin{tabular}{lccc}
  \toprule
  \multirow{1.1}{*}{\bfseries{Correspondence}}& \multicolumn{3}{c}{\bfseries{Top K}}\\
   \hhline{~---}  
  \multirow{1.25}{*}{\bfseries{Methods}}& 1 & 3 & 5 \\ 
  \hline
  \hline
   \multicolumn{1}{l|}{DIFT~\cite{tang2023dift}}& 54.5 & 58.2 & \textbf{60.7}\\
   \multicolumn{1}{l|}{SD-DINO~\cite{zhang2023tale}}& 53.5& 59.3& 58.2\\
   \multicolumn{1}{l|}{DINO-VIT~\cite{carion2020end}}& 37.5& 32.4& 36.4\\
  \bottomrule
  \end{tabular}}
    \vspace{-0.3cm}
\end{table}
\section{Conclusion}
%----------------------------------------------------------------------
Our work focuses on enabling robotic manipulation to generalize beyond object categories, which is crucial for embodied intelligence toward the open world. We tackle the challenge of learning to interact with various objects and transferring knowledge across different categories. Inspired by the cognition process of humans, we extract an “affordance memory” containing diverse object interaction information from human videos, then retrieve relevant objects from the memory based on visual and semantic similarity. Combined with a powerful diffusion model-based semantic correspondence mapping, our approach achieves significant generalization ability using only a small-scale memory information. Notably, our method achieves unsupervised zero-shot generalization without manual annotation, additional training, or human prior. We hope that our work will inspire future research in this direction and contribute to the development of embodied intelligence in the open world.
\section*{Acknowledgement}
\vspace{-0.2cm}
This work is supported by the National Key R\&D Program of China (2022ZD0161700).
%-------------------------------------------------------------------
%%%%%%%%%%%%%%%%%%%%%%%%%%%%%%%%%%%%%%%%%%%%%%%%%%%%%%%%%%%%%%%%%%%%%%%%%%%%%%%%

\clearpage
\newpage
% \mbox{~}
% \clearpage
% \newpage
{\small
\bibliographystyle{ieee_fullname}
\bibliography{egbib}
}
\clearpage
\appendix
\section*{Appendix}
\input{supp/supp}
\end{document}

%% file: supp/supp.tex
\section{Introduction on Baselines}
\subsection{End-to-end Affordance Model}
\label{sec:affordance_model}
The main baselines we compare in this paper are end-to-end affordance prediction methods, including HOI\cite{liu2022joint}, HAP\cite{goyal2022human}, HOTSPOTS\cite{nagarajan2019grounded}, VRB\cite{bahl2023affordances}, and LOCATE\cite{luo2022learning}. We will briefly introduce these baselines: HOI\cite{liu2022joint} primarily predicts the trajectory of the hand and the area of the object that will be contacted in the future. HAP\cite{goyal2022human} focuses on observing the state of the hand to learn state-sensitive features and object affordances, including interaction region and the grasping pose. HOTSPOTS\cite{nagarajan2019grounded} propose to perform affordance grounding, which involves determining where on the current object can the interaction occur given a specific action. VRB\cite{bahl2023affordances} models an object's affordance as the contact area and the subsequent motion trajectory,then tries to predict them. LOCATE\cite{luo2022learning} focuses on images of human-object interaction to achieve affordance grounding and can transfer parts across different categories given a source image. The need for the source image makes it a "one-shot" method in contrast to our zero-shot approach. %This work also introduces the evaluation dataset that we utilize.

Apart from LOCATE \cite{luo2022learning}, all other baseline methods output heatmaps concerning the contact location. Particularly, both VRB\cite{bahl2023affordances} and HOI\cite{liu2022joint} predict both the contact area and the trajectory of the hand's movement. However, in this work, we solely focus on the object's contact area and do not consider the trajectory after the contact. For each baseline, we directly use the available pre-trained models that were trained on EPIC-KITCHEN\cite{damen2020epic}. In the main text, we argue that for robot manipulation, the contact should be on points rather than regions. Therefore, for all heatmaps output from the baseline methods, we selected the top five points in the heatmap for subsequent evaluation. 

When evaluating LOCATE\cite{luo2022learning},  one label is needed for the action to be performed. For the objects selected from the original AGD-20K dataset, since they all belong to the “hold” action category of the seen setting, we organize these objects into this specific category. As for the newly collected door and drawer categories, we place them within the “open” action category of the unseen setting. For each input action label and corresponding image, LOCATE generates a localization map composed of normalized activation values, which serves as a representation of predictions for the affordance region. Consequently, we select the top five points with the highest activation values for comparison.

% dataset,top5
\subsection{Semantic Correspondence Methods}
\label{sec:semantic_correspondence_app}
% explain what's dift, sd-dino, dino-vit, ldm-sc
Semantic correspondence maps pixel location in the source image to the corresponding pixel location in the target image, such that the pair of pixels bear a similar semantic meaning. These methods can find the semantic correspondence beyond the category of objects, like the wings of a bird and the wings of an airplane, Recent advances find that foundation visual models have the ability to find semantic correspondence in a zero-shot manner, without additional training or finetuning.

We select two lines of works in zero-shot semantic correspondence as baselines, one is based on the idea of dense feature matching, and the other is based on special token optimization. The former one includes DIFT\cite{tang2023dift}, SD-DINO\cite{zhang2023tale}, and DINO-ViT\cite{carion2020end}, while the latter one includes LDM-SC\cite{hedlin2023unsupervised}. We use the official code of these methods to get the semantic correspondence results. 

Methods based on feature mapping first extract the feature map of the source image and the target image via encoders of the visual model, then match the feature of the source pixel to its nearest neighbor in the target image. The matching is done by calculating the cosine similarity between the source feature and the target feature. Such a straightforward approach has proven to be effective in both DINO features~\cite{carion2020end} and DIFT features~\cite{tang2023dift}. And SD-DINO~\cite{zhang2023tale} seeks to merge DINO and DIFT features to get better performance. These methods are easy to implement and don't need training or optimization of any kind, thus having a very fast inference speed (about 30s on an A40 GPU).

Methods based on special token optimization like LDM-SC~\cite{hedlin2023unsupervised} take a different approach. They try to find a special token in the language latent space of a text-to-image diffusion model that ``describes'' the semantic meaning of the source pixel, then find the pixel that matches the special token best in the target model. More specifically, they first calculate the cross-attention map between the special token and the source image, and optimize the special token to maximize the attention value at the source pixel. Then, they calculate the cross-attention map between the optimized special token and the target image, and identify the target pixel as the one with the highest attention valueassociated with the special token. Because each mapping needs to optimize a special token, this approach is slower than dense feature mapping methods (about 5 min on an A40 GPU).

We select DIFT as the semantic correspondence method in \method, since it has the most balanced performance in accuracy and inference speed. Please refer to Section \ref{sec:retriever_correspondence} for a visualized comparison between these methods.

\section{Implementation Details}
\subsection{Selection of Affordance Buffer}

We extract affordance information from the EpicKitchens-100 Video~\cite{damen2020epic} dataset. We visualize all the videos, filtering out and removing those with poor lighting conditions, low video clarity, or the constant presence of objects obstructing the view. The extraction process is similar to that of VRB\cite{bahl2023affordances} and HOI\cite{liu2022joint}. In addition, we added a window for motion blur detection. In this process, we employ the Laplacian operator to detect and quantify motion blur, ensuring that we can crop and select the clearest frame near the frame where contact happens.  We use skin segmentation ~\cite{saxen2014color} to detect the intersection of the bounding box (bbox) of the object $B_o$ and the hand $B_h$. We then randomly sample contact points within this area. Because some objects have a small surface area, and due to the errors in the mapping of the homography matrix between different frames, the sampled points may be outside the object. This is a significant challenge when it comes to semantic correspondence. To solve this problem, we took the average position of all the points that were mapped back. For most objects, the average position will be on the object. Then, we take this average position and randomly sample five points within a circle with a radius of four pixels around it. This procedure ensures that all the sampled contact points are on the object. The object categories we extracted are shown in the table below:

\begin{table}[ht]
\centering
\caption{List of all object categories in the affordance memory}
\label{tab:objects}
\begin{tabularx}{.48\textwidth}{@{}>{\centering\arraybackslash}m{1.5cm}X@{}}
\toprule
\textbf{All \newline categories}   & bottle, bowl, drawer, cup, door, fork, knife, cupboard, scissors, wine glass, banana, bread machine, tap, pot, lid, plate, fridge, bag, trash bin, kettle, oven, pizza, mug, cucumber, peeler, mouse, rice cooker, bag, spatula, slicer, computer keyboard, phone, container, hob, heater, onion, tray, melon, coffee maker, remote, dishwasher, spoon, processor, sponge,  package, dough, meat, cheese, blender, button, tomato\\
\bottomrule
\end{tabularx}
\end{table}

\subsection{Evaluation Dataset}
Apart from the evaluation dataset from LOCATE, we collect two new types of objects using Labelme: door and drawer. For a cabinet, there may be multiple handles. When we mark the Ground Truth (GT), we mark all the possible interactive positions to ensure the fairness of the validation.

\begin{figure}[ht]
    \centering
    % \vspace{-0.1cm}
    \includegraphics[width=0.48\textwidth]{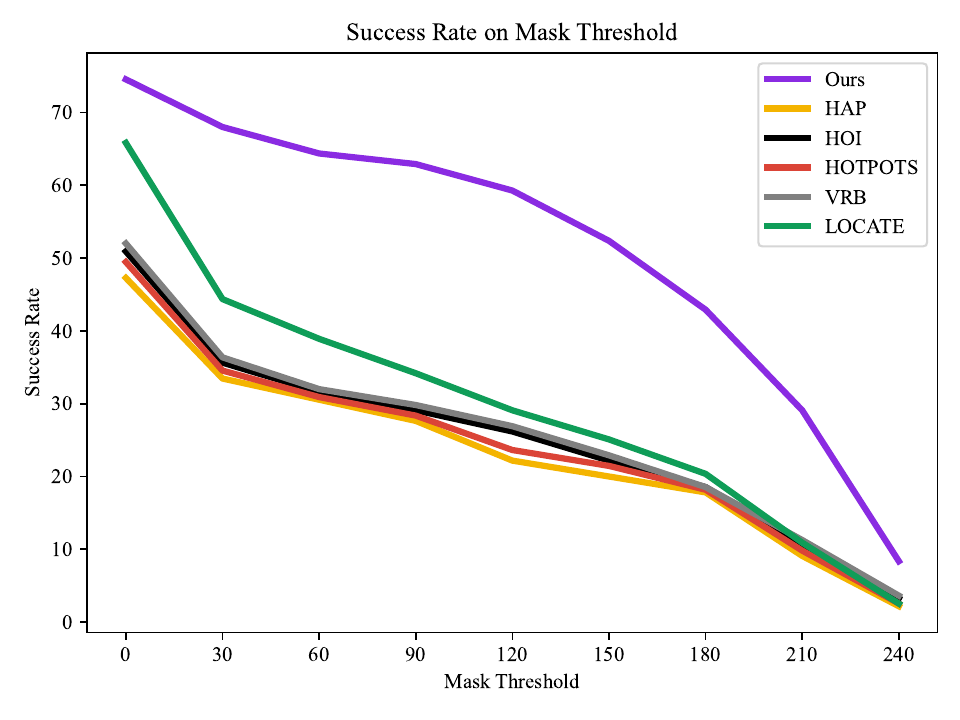}
    \vspace{-0.5cm}
    \caption{\textbf{Success rate on different mask threshold.} \method always exceeds other baselines by a large margin.
    }
    % \vspace{-0.1cm}
    \label{fig:sr_curve}
\end{figure}

% \subsection{Evaluation Metrics}
% For normalized scanpath saliency (NSS), we not only apply an inverse version of the metric such that we are calculating the mask values of the corresponding affordance outputs on the ground truth mask but also skip the normalization step in the initial formula of NSS. We find that the unnormalized ground truth mask has a better alignment property on different objects, such that the roughly same threshold value of 60 matches the real boundary of most objects from different categories.

\subsection{Memory Retrieval}
In the process of memory retrieval through visual and semantic similarity, we utilize different encoders to test the effectiveness of the retrieval results. Notably, we compare CLIP-B32, CLIP-B16 and LPIPS metrics. 

For the CLIP-B32+LPIPS retriever, we use the CLIP-B32 encoder to retrieve the five images from the affordance memory that are most semantically similar. Subsequently, we employ a pre-trained VGG network to identify the one with the minimum LPIPS value among these five images, signifying the visually most similar image to the source.
\subsection{Affordance Mapping}
\label{sec:affordance_mapping}
During the extraction of affordance memory, we retrieved more than one contact point on the object, and experimented with two different ways of mapping the affordance memory to the target image. The first one is to map each source point to the target image separately, and then average the results. The second one is to first average the contact points, then map the average point to the target image. We found that while the second method is more effective, the first approach yields slightly better results, and we use this method in our experiments.
% Average after mapping / before mapping
% A small table comparing these two methods
\subsection{Robot Hardware Setup}
\label{app:robot_hardware}
For the real-world robot setup, we use the Franka Panda robot, and we set up a calibrated RealSense L515 camera externally on the robot for observations.

%action space：
\begin{figure}[h]
    \centering
    % \vspace{-0.1cm}
    \includegraphics[width=0.48\textwidth]{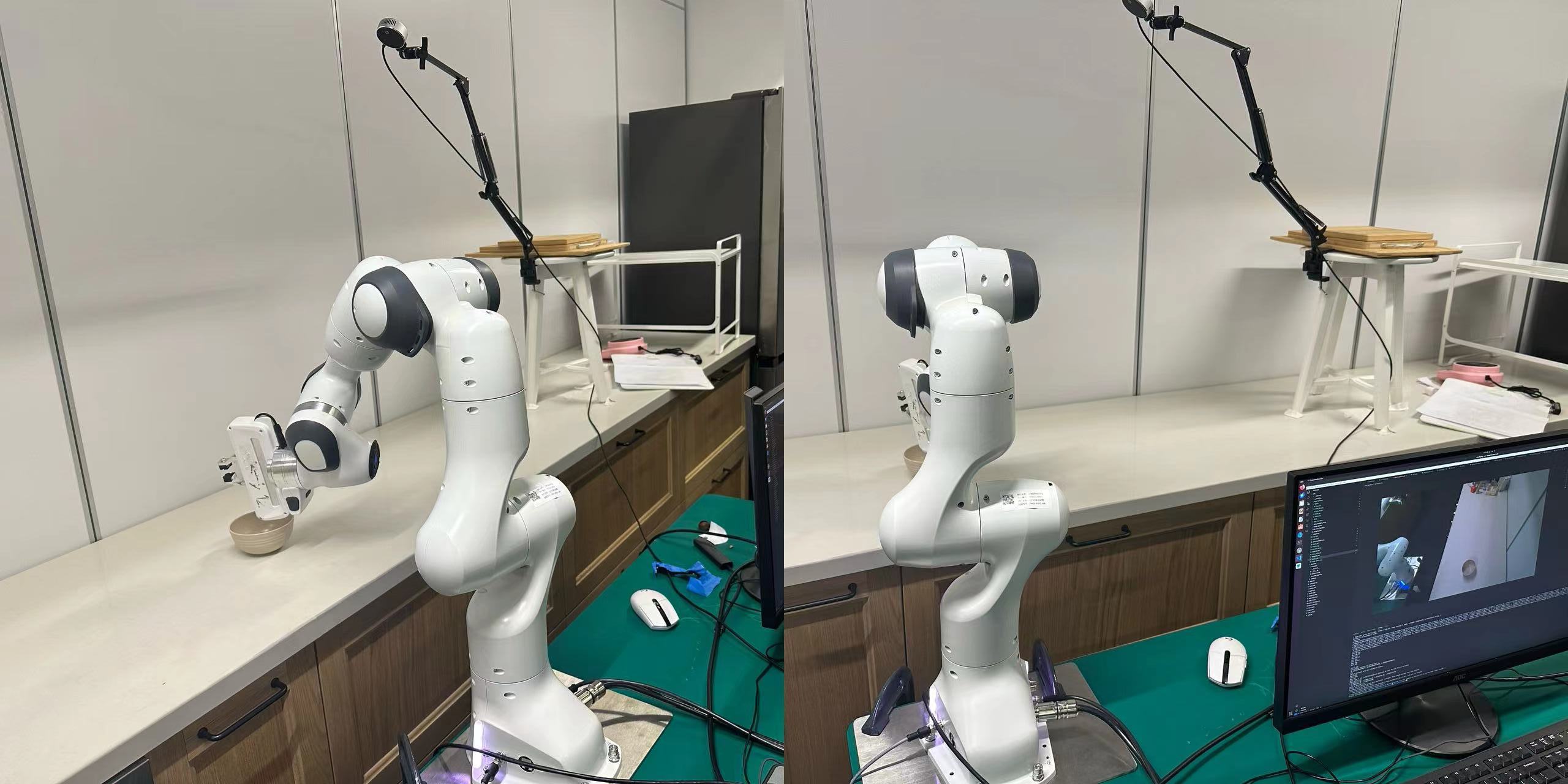}
    \vspace{-0.5cm}
    \caption{\textbf{Our workspace.}
    }
    % \vspace{-0.1cm}
    \label{fig:main_workspace}
\end{figure}
\begin{figure*}[t]
    \centering
    % \vspace{-0.1cm}
    \includegraphics[width=1.0\textwidth]{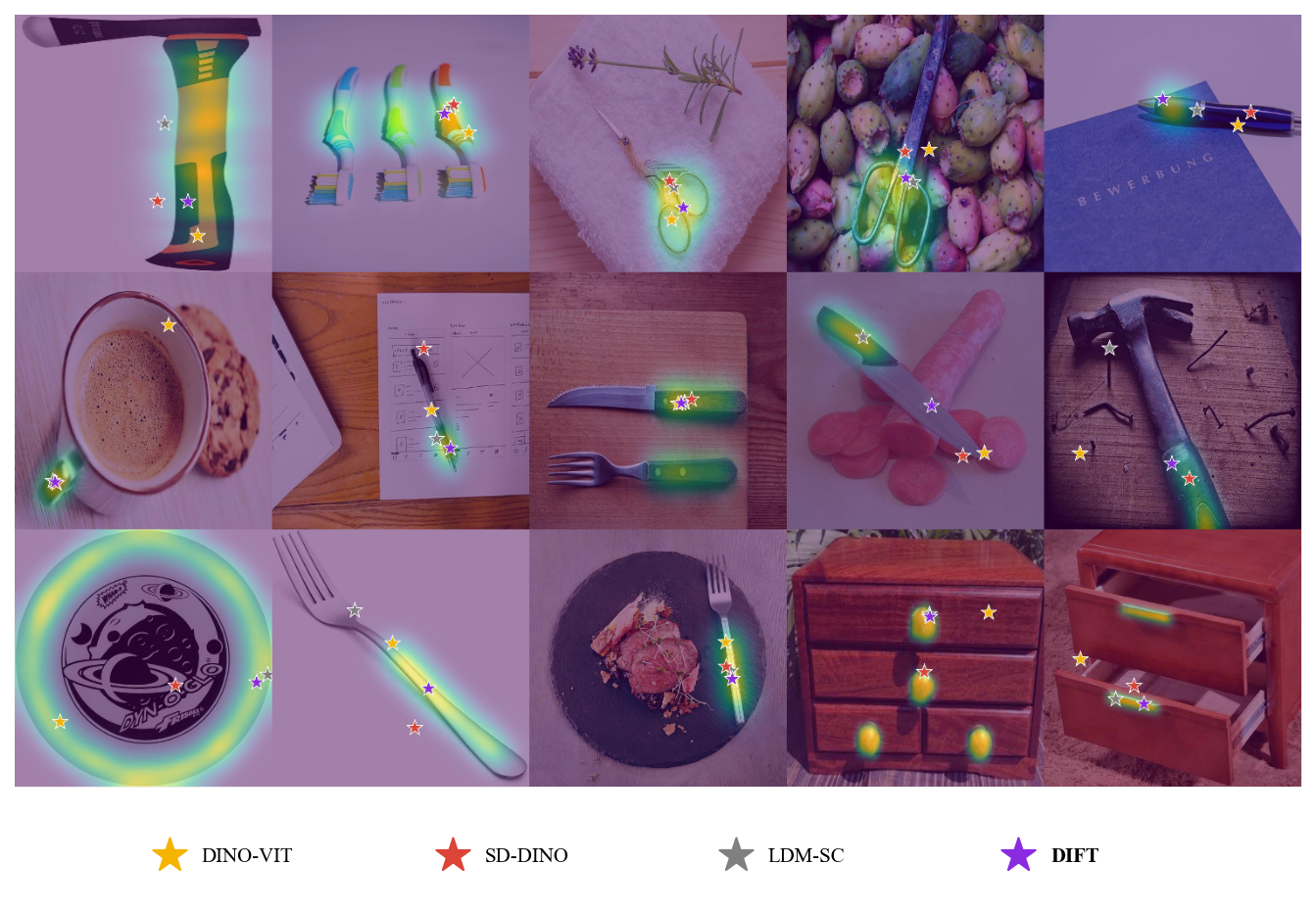}
    \vspace{-0.5cm}
    \caption{\textbf{Visualization of different correspondence methods on the same source image.}
    }
    % \vspace{-0.1cm}
    \label{fig:correspondence_ablation}
\end{figure*}
% camera, robot, calibration, workspace photo
\section{Additional Results}

% \subsection{Cross Category Baselines Performance}
% % Baseline performance on Figure 3
% In this section, we present visualized results of other end-to-end baselines in Fig\ref{fig:all_types} for comparison. These results clearly reveal the superior performance of our approach: even when the number of object classes that have been seen is limited, Robo-ABC still performs remarkably well in zero-shot learning, demonstrating strong cross-category generalization.

\subsection{Correspondence}
\label{sec:retriever_correspondence}
We visualize the correspondence results for different correspondence models and show them in Figure \ref{fig:correspondence_ablation}. DIFT-based methods yield the best correspondence results while having a faster inference speed.
\subsection{Real Robot Results}
\label{app:robot_results}
The videos of real-world robot deployment results can be found in the supplementary material. We recommend watching the videos to have a better sense of \method's ability. We report the success rate of our method for seven object categories along with VRB\cite{bahl2023affordances} in Table \ref{tab:real_world}. For each experiment, we reposition the object and try to grasp for five times and report the overall average success rate.

\begin{table}[!ht]
\centering 
\renewcommand{\arraystretch}{1.2}
\begin{tabularx}{0.48\textwidth}{l *{4}{>{\centering\arraybackslash}X}}
\toprule
\textbf{Object} & bowl & bottle & racket & scissor \\ \midrule
VRB & 3/5 & 5/5 & 2/5 & 4/5 \\ 
Ours & 4/5 & 5/5 & 5/5 & 4/5 \\ \bottomrule
\end{tabularx}
\newline
\vspace*{-0.5 cm} % Adds vertical space between the tables
\end{table}
% \newline

\begin{table}[!ht]
\centering 
\renewcommand{\arraystretch}{1.2}
\begin{tabularx}{0.48\textwidth}{l *{4}{>{\centering\arraybackslash}X}}
\toprule 
\textbf{Object} & knife & cup & glass & \bfseries{overall} \\
\midrule 
VRB  & 2/5 & 3/5 & 5/5 & 68.6\% \\
\bfseries{Ours} & 3/5 & 5/5 & 4/5 & \bfseries{85.7\%} \\
\bottomrule 
\end{tabularx}
\caption{Real-World Success Rate} 
\label{tab:real_world} 
\end{table}

\section{Limitations}
Although Robo-ABC has shown significant improvement in output accuracy compared to previous end-to-end methods and no manual annotation  requirement or additional training, there are still many directions for improvement. In the process of extracting object affordances from human videos, we need to visualize all outputs to check the usability of the affordances produced. Moreover, the low resolution of the original dataset and the unusability of some videos due to lighting problems can affect the accuracy of finding semantic correspondences in the subsequent stage. 
% Also, the output of \method is in the 2D space and may be insufficient to provide guidance for dexterous operation.
% Per category success rate and video

% \begin{table}[ht]
% \centering
% \caption{List of seen and unseen objects for affordance memory}
% \label{tab:objects}
% \begin{tabularx}{.5\textwidth}{@{}>{\centering\arraybackslash}m{1.5cm}X@{}}
% \toprule
% \textbf{All categories}   & bottle, bowl, drawer, cup, door, \\
%                 & fork, knife, scissors, wine glass \\
% \bottomrule
% \end{tabularx}
% \label{tab:seen_unseen_objects}
% \end{table}

%% file: output.bbl
\begin{thebibliography}{10}
\providecommand{\url}[1]{#1}
\csname url@samestyle\endcsname
\providecommand{\newblock}{\relax}
\providecommand{\bibinfo}[2]{#2}
\providecommand{\BIBentrySTDinterwordspacing}{\spaceskip=0pt\relax}
\providecommand{\BIBentryALTinterwordstretchfactor}{4}
\providecommand{\BIBentryALTinterwordspacing}{\spaceskip=\fontdimen2\font plus
\BIBentryALTinterwordstretchfactor\fontdimen3\font minus \fontdimen4\font\relax}
\providecommand{\BIBforeignlanguage}[2]{{%
\expandafter\ifx\csname l@#1\endcsname\relax
\typeout{** WARNING: IEEEtran.bst: No hyphenation pattern has been}%
\typeout{** loaded for the language `#1'. Using the pattern for}%
\typeout{** the default language instead.}%
\else
\language=\csname l@#1\endcsname
\fi
#2}}
\providecommand{\BIBdecl}{\relax}
\BIBdecl

\bibitem{carion2020end}
N.~Carion, F.~Massa, G.~Synnaeve, N.~Usunier, A.~Kirillov, and S.~Zagoruyko, ``End-to-end object detection with transformers,'' in \emph{European conference on computer vision}.\hskip 1em plus 0.5em minus 0.4em\relax Springer, 2020, pp. 213--229.

\bibitem{zhu2014reasoning}
Y.~Zhu, A.~Fathi, and L.~Fei-Fei, ``Reasoning about object affordances in a knowledge base representation,'' in \emph{Computer Vision--ECCV 2014: 13th European Conference, Zurich, Switzerland, September 6-12, 2014, Proceedings, Part II 13}.\hskip 1em plus 0.5em minus 0.4em\relax Springer, 2014, pp. 408--424.

\bibitem{gibson1978ecological}
J.~J. Gibson, ``The ecological approach to the visual perception of pictures,'' \emph{Leonardo}, vol.~11, no.~3, pp. 227--235, 1978.

\bibitem{myers2015affordance}
A.~Myers, C.~L. Teo, C.~Ferm{\"u}ller, and Y.~Aloimonos, ``Affordance detection of tool parts from geometric features,'' in \emph{2015 IEEE International Conference on Robotics and Automation (ICRA)}.\hskip 1em plus 0.5em minus 0.4em\relax IEEE, 2015, pp. 1374--1381.

\bibitem{amir2021deep}
S.~Amir, Y.~Gandelsman, S.~Bagon, and T.~Dekel, ``Deep vit features as dense visual descriptors,'' \emph{arXiv preprint arXiv:2112.05814}, vol.~2, no.~3, p.~4, 2021.

\bibitem{jiang2023doduo}
Z.~Jiang, H.~Jiang, and Y.~Zhu, ``Doduo: Dense visual correspondence from unsupervised semantic-aware flow,'' in \emph{arXiv preprint arXiv:2309.15110}, 2023.

\bibitem{tang2023emergent}
L.~Tang, M.~Jia, Q.~Wang, C.~P. Phoo, and B.~Hariharan, ``Emergent correspondence from image diffusion,'' \emph{arXiv preprint arXiv:2306.03881}, 2023.

\bibitem{hedlin2023unsupervised}
E.~Hedlin, G.~Sharma, S.~Mahajan, H.~Isack, A.~Kar, A.~Tagliasacchi, and K.~M. Yi, ``Unsupervised semantic correspondence using stable diffusion,'' \emph{arXiv preprint arXiv:2305.15581}, 2023.

\bibitem{nair2022r3m}
S.~Nair, A.~Rajeswaran, V.~Kumar, C.~Finn, and A.~Gupta, ``R3m: A universal visual representation for robot manipulation,'' \emph{arXiv preprint arXiv:2203.12601}, 2022.

\bibitem{shan2020understanding}
D.~Shan, J.~Geng, M.~Shu, and D.~F. Fouhey, ``Understanding human hands in contact at internet scale,'' in \emph{Proceedings of the IEEE/CVF conference on computer vision and pattern recognition}, 2020, pp. 9869--9878.

\bibitem{bahl2023affordances}
S.~Bahl, R.~Mendonca, L.~Chen, U.~Jain, and D.~Pathak, ``Affordances from human videos as a versatile representation for robotics,'' in \emph{Proceedings of the IEEE/CVF Conference on Computer Vision and Pattern Recognition}, 2023, pp. 13\,778--13\,790.

\bibitem{goyal2022human}
M.~Goyal, S.~Modi, R.~Goyal, and S.~Gupta, ``Human hands as probes for interactive object understanding,'' in \emph{Computer Vision and Pattern Recognition (CVPR)}, 2022.

\bibitem{fang2023anygrasp}
H.-S. Fang, C.~Wang, H.~Fang, M.~Gou, J.~Liu, H.~Yan, W.~Liu, Y.~Xie, and C.~Lu, ``Anygrasp: Robust and efficient grasp perception in spatial and temporal domains,'' \emph{IEEE Transactions on Robotics}, 2023.

\bibitem{brohan2023can}
A.~Brohan, Y.~Chebotar, C.~Finn, K.~Hausman, A.~Herzog, D.~Ho, J.~Ibarz, A.~Irpan, E.~Jang, R.~Julian \emph{et~al.}, ``Do as i can, not as i say: Grounding language in robotic affordances,'' in \emph{Conference on Robot Learning}.\hskip 1em plus 0.5em minus 0.4em\relax PMLR, 2023, pp. 287--318.

\bibitem{wang2023d}
Y.~Wang, Z.~Li, M.~Zhang, K.~Driggs-Campbell, J.~Wu, L.~Fei-Fei, and Y.~Li, ``$d^3$ fields: Dynamic 3d descriptor fields for zero-shot generalizable robotic manipulation,'' \emph{arXiv preprint arXiv:2309.16118}, 2023.

\bibitem{xue2023useek}
Z.~Xue, Z.~Yuan, J.~Wang, X.~Wang, Y.~Gao, and H.~Xu, ``Useek: Unsupervised se (3)-equivariant 3d keypoints for generalizable manipulation,'' in \emph{2023 IEEE International Conference on Robotics and Automation (ICRA)}.\hskip 1em plus 0.5em minus 0.4em\relax IEEE, 2023, pp. 1715--1722.

\bibitem{geng2023gapartnet}
H.~Geng, H.~Xu, C.~Zhao, C.~Xu, L.~Yi, S.~Huang, and H.~Wang, ``Gapartnet: Cross-category domain-generalizable object perception and manipulation via generalizable and actionable parts,'' in \emph{Proceedings of the IEEE/CVF Conference on Computer Vision and Pattern Recognition}, 2023, pp. 7081--7091.

\bibitem{geng2023partmanip}
H.~Geng, Z.~Li, Y.~Geng, J.~Chen, H.~Dong, and H.~Wang, ``Partmanip: Learning cross-category generalizable part manipulation policy from point cloud observations,'' in \emph{Proceedings of the IEEE/CVF Conference on Computer Vision and Pattern Recognition}, 2023, pp. 2978--2988.

\bibitem{geng2023rlafford}
Y.~Geng, B.~An, H.~Geng, Y.~Chen, Y.~Yang, and H.~Dong, ``Rlafford: End-to-end affordance learning for robotic manipulation,'' in \emph{2023 IEEE International Conference on Robotics and Automation (ICRA)}.\hskip 1em plus 0.5em minus 0.4em\relax IEEE, 2023, pp. 5880--5886.

\bibitem{ning2023where2explore}
C.~Ning, R.~Wu, H.~Lu, K.~Mo, and H.~Dong, ``Where2explore: Few-shot affordance learning for unseen novel categories of articulated objects,'' \emph{arXiv preprint arXiv:2309.07473}, 2023.

\bibitem{cheng2023learning}
K.~Cheng, R.~Wu, Y.~Shen, C.~Ning, G.~Zhan, and H.~Dong, ``Learning environment-aware affordance for 3d articulated object manipulation under occlusions,'' \emph{arXiv preprint arXiv:2309.07510}, 2023.

\bibitem{wang2022adaafford}
Y.~Wang, R.~Wu, K.~Mo, J.~Ke, Q.~Fan, L.~J. Guibas, and H.~Dong, ``Adaafford: Learning to adapt manipulation affordance for 3d articulated objects via few-shot interactions,'' in \emph{European Conference on Computer Vision}.\hskip 1em plus 0.5em minus 0.4em\relax Springer, 2022, pp. 90--107.

\bibitem{wu2021vat}
R.~Wu, Y.~Zhao, K.~Mo, Z.~Guo, Y.~Wang, T.~Wu, Q.~Fan, X.~Chen, L.~Guibas, and H.~Dong, ``Vat-mart: Learning visual action trajectory proposals for manipulating 3d articulated objects,'' \emph{arXiv preprint arXiv:2106.14440}, 2021.

\bibitem{mo2022o2o}
K.~Mo, Y.~Qin, F.~Xiang, H.~Su, and L.~Guibas, ``O2o-afford: Annotation-free large-scale object-object affordance learning,'' in \emph{Conference on Robot Learning}.\hskip 1em plus 0.5em minus 0.4em\relax PMLR, 2022, pp. 1666--1677.

\bibitem{mo2021where2act}
K.~Mo, L.~J. Guibas, M.~Mukadam, A.~Gupta, and S.~Tulsiani, ``Where2act: From pixels to actions for articulated 3d objects,'' in \emph{Proceedings of the IEEE/CVF International Conference on Computer Vision}, 2021, pp. 6813--6823.

\bibitem{wu2023learning}
Y.-H. Wu, J.~Wang, and X.~Wang, ``Learning generalizable dexterous manipulation from human grasp affordance,'' in \emph{Conference on Robot Learning}.\hskip 1em plus 0.5em minus 0.4em\relax PMLR, 2023, pp. 618--629.

\bibitem{kannan2023deft}
A.~Kannan, K.~Shaw, S.~Bahl, P.~Mannam, and D.~Pathak, ``Deft: Dexterous fine-tuning for real-world hand policies,'' \emph{arXiv preprint arXiv:2310.19797}, 2023.

\bibitem{nagarajan2019grounded}
T.~Nagarajan, C.~Feichtenhofer, and K.~Grauman, ``Grounded human-object interaction hotspots from video,'' in \emph{Proceedings of the IEEE/CVF International Conference on Computer Vision}, 2019, pp. 8688--8697.

\bibitem{liu2022joint}
S.~Liu, S.~Tripathi, S.~Majumdar, and X.~Wang, ``Joint hand motion and interaction hotspots prediction from egocentric videos,'' in \emph{Proceedings of the IEEE/CVF Conference on Computer Vision and Pattern Recognition}, 2022, pp. 3282--3292.

\bibitem{luo2022learning}
H.~Luo, W.~Zhai, J.~Zhang, Y.~Cao, and D.~Tao, ``Learning affordance grounding from exocentric images,'' in \emph{Proceedings of the IEEE/CVF Conference on Computer Vision and Pattern Recognition}, 2022, pp. 2252--2261.

\bibitem{ye2023affordance}
Y.~Ye, X.~Li, A.~Gupta, S.~De~Mello, S.~Birchfield, J.~Song, S.~Tulsiani, and S.~Liu, ``Affordance diffusion: Synthesizing hand-object interactions,'' in \emph{Proceedings of the IEEE/CVF Conference on Computer Vision and Pattern Recognition}, 2023, pp. 22\,479--22\,489.

\bibitem{hou2021affordance}
Z.~Hou, B.~Yu, Y.~Qiao, X.~Peng, and D.~Tao, ``Affordance transfer learning for human-object interaction detection,'' in \emph{Proceedings of the IEEE/CVF Conference on Computer Vision and Pattern Recognition}, 2021, pp. 495--504.

\bibitem{hassan2016attribute}
M.~Hassan and A.~Dharmaratne, ``Attribute based affordance detection from human-object interaction images,'' in \emph{Image and Video Technology--PSIVT 2015 Workshops: RV 2015, GPID 2013, VG 2015, EO4AS 2015, MCBMIIA 2015, and VSWS 2015, Auckland, New Zealand, November 23-27, 2015. Revised Selected Papers 7}.\hskip 1em plus 0.5em minus 0.4em\relax Springer, 2016, pp. 220--232.

\bibitem{damen2022rescaling}
D.~Damen, H.~Doughty, G.~M. Farinella, A.~Furnari, E.~Kazakos, J.~Ma, D.~Moltisanti, J.~Munro, T.~Perrett, W.~Price \emph{et~al.}, ``Rescaling egocentric vision: Collection, pipeline and challenges for epic-kitchens-100,'' \emph{International Journal of Computer Vision}, pp. 1--23, 2022.

\bibitem{damen2018scaling}
D.~Damen, H.~Doughty, G.~M. Farinella, S.~Fidler, A.~Furnari, E.~Kazakos, D.~Moltisanti, J.~Munro, T.~Perrett, W.~Price \emph{et~al.}, ``Scaling egocentric vision: The epic-kitchens dataset,'' in \emph{Proceedings of the European conference on computer vision (ECCV)}, 2018, pp. 720--736.

\bibitem{damen2020epic}
------, ``The epic-kitchens dataset: Collection, challenges and baselines,'' \emph{IEEE Transactions on Pattern Analysis and Machine Intelligence}, vol.~43, no.~11, pp. 4125--4141, 2020.

\bibitem{grauman2022ego4d}
K.~Grauman, A.~Westbury, E.~Byrne, Z.~Chavis, A.~Furnari, R.~Girdhar, J.~Hamburger, H.~Jiang, M.~Liu, X.~Liu \emph{et~al.}, ``Ego4d: Around the world in 3,000 hours of egocentric video,'' in \emph{Proceedings of the IEEE/CVF Conference on Computer Vision and Pattern Recognition}, 2022, pp. 18\,995--19\,012.

\bibitem{mendonca23swim}
R.~Mendonca, S.~Bahl, and D.~Pathak, ``Structured world models from human videos,'' \emph{RSS}, 2023.

\bibitem{bahl2022human}
S.~Bahl, A.~Gupta, and D.~Pathak, ``Human-to-robot imitation in the wild,'' \emph{RSS}, 2022.

\bibitem{zhang2023tale}
J.~Zhang, C.~Herrmann, J.~Hur, L.~P. Cabrera, V.~Jampani, D.~Sun, and M.-H. Yang, ``A tale of two features: Stable diffusion complements dino for zero-shot semantic correspondence,'' \emph{arXiv preprint arXiv:2305.15347}, 2023.

\bibitem{demo2vec2018cvpr}
K.~Fang, T.-L. Wu, D.~Yang, S.~Savarese, and J.~J. Lim, ``Demo2vec: Reasoning object affordances from online videos,'' in \emph{The IEEE Conference on Computer Vision and Pattern Recognition (CVPR)}, June 2018.

\bibitem{saxen2014color}
F.~Saxen and A.~Al-Hamadi, ``Color-based skin segmentation: An evaluation of the state of the art,'' in \emph{2014 IEEE International Conference on Image Processing (ICIP)}.\hskip 1em plus 0.5em minus 0.4em\relax IEEE, 2014, pp. 4467--4471.

\bibitem{creem2005neural}
S.~H. Creem-Regehr and J.~N. Lee, ``Neural representations of graspable objects: are tools special?'' \emph{Cognitive Brain Research}, vol.~22, no.~3, pp. 457--469, 2005.

\bibitem{tang2023dift}
L.~Tang, M.~Jia, Q.~Wang, C.~P. Phoo, and B.~Hariharan, ``Emergent correspondence from image diffusion,'' \emph{arXiv preprint arXiv:2306.03881}, 2023.

\bibitem{luca2023lang}
L.~Medeiros, ``lang-segment-anything,'' \url{https://github.com/luca-medeiros/lang-segment-anything}, 2023.

\bibitem{Radford2021LearningTV}
A.~Radford, J.~W. Kim, C.~Hallacy, A.~Ramesh, G.~Goh, S.~Agarwal, G.~Sastry, A.~Askell, P.~Mishkin, J.~Clark, G.~Krueger, and I.~Sutskever, ``Learning transferable visual models from natural language supervision,'' in \emph{International Conference on Machine Learning}, 2021.

\bibitem{xu2023unidexgrasp}
Y.~Xu, W.~Wan, J.~Zhang, H.~Liu, Z.~Shan, H.~Shen, R.~Wang, H.~Geng, Y.~Weng, J.~Chen \emph{et~al.}, ``Unidexgrasp: Universal robotic dexterous grasping via learning diverse proposal generation and goal-conditioned policy,'' in \emph{Proceedings of the IEEE/CVF Conference on Computer Vision and Pattern Recognition}, 2023, pp. 4737--4746.

\bibitem{wan2023unidexgrasp++}
W.~Wan, H.~Geng, Y.~Liu, Z.~Shan, Y.~Yang, L.~Yi, and H.~Wang, ``Unidexgrasp++: Improving dexterous grasping policy learning via geometry-aware curriculum and iterative generalist-specialist learning,'' \emph{arXiv preprint arXiv:2304.00464}, 2023.

\bibitem{rashid2023language}
A.~Rashid, S.~Sharma, C.~M. Kim, J.~Kerr, L.~Y. Chen, A.~Kanazawa, and K.~Goldberg, ``Language embedded radiance fields for zero-shot task-oriented grasping,'' in \emph{7th Annual Conference on Robot Learning}, 2023.

\bibitem{mandikal2021learning}
P.~Mandikal and K.~Grauman, ``Learning dexterous grasping with object-centric visual affordances,'' in \emph{2021 IEEE international conference on robotics and automation (ICRA)}.\hskip 1em plus 0.5em minus 0.4em\relax IEEE, 2021, pp. 6169--6176.

\bibitem{mandikal2022dexvip}
------, ``Dexvip: Learning dexterous grasping with human hand pose priors from video,'' in \emph{Conference on Robot Learning}.\hskip 1em plus 0.5em minus 0.4em\relax PMLR, 2022, pp. 651--661.

\bibitem{li2023gendexgrasp}
P.~Li, T.~Liu, Y.~Li, Y.~Geng, Y.~Zhu, Y.~Yang, and S.~Huang, ``Gendexgrasp: Generalizable dexterous grasping,'' in \emph{2023 IEEE International Conference on Robotics and Automation (ICRA)}.\hskip 1em plus 0.5em minus 0.4em\relax IEEE, 2023, pp. 8068--8074.

\bibitem{reed2022generalist}
S.~Reed, K.~Zolna, E.~Parisotto, S.~G. Colmenarejo, A.~Novikov, G.~Barth-Maron, M.~Gimenez, Y.~Sulsky, J.~Kay, J.~T. Springenberg \emph{et~al.}, ``A generalist agent,'' \emph{arXiv preprint arXiv:2205.06175}, 2022.

\bibitem{zhang2018perceptual}
R.~Zhang, P.~Isola, A.~A. Efros, E.~Shechtman, and O.~Wang, ``The unreasonable effectiveness of deep features as a perceptual metric,'' in \emph{CVPR}, 2018.

\bibitem{luo2023dhf}
G.~Luo, L.~Dunlap, D.~H. Park, A.~Holynski, and T.~Darrell, ``Diffusion hyperfeatures: Searching through time and space for semantic correspondence,'' in \emph{Advances in Neural Information Processing Systems}, 2023.

\bibitem{ye2023foundations}
W.~Ye, Y.~Zhang, M.~Wang, S.~Wang, X.~Gu, P.~Abbeel, and Y.~Gao, ``Foundation reinforcement learning: towards embodied generalist agents with foundation prior assistance,'' \emph{arXiv preprint arXiv:2310.02635}, 2023.

\bibitem{gao2023can}
J.~Gao, K.~Hu, G.~Xu, and H.~Xu, ``Can pre-trained text-to-image models generate visual goals for reinforcement learning?'' \emph{arXiv preprint arXiv:2307.07837}, 2023.

\bibitem{radford2021learning}
A.~Radford, J.~W. Kim, C.~Hallacy, A.~Ramesh, G.~Goh, S.~Agarwal, G.~Sastry, A.~Askell, P.~Mishkin, J.~Clark \emph{et~al.}, ``Learning transferable visual models from natural language supervision,'' in \emph{International conference on machine learning}.\hskip 1em plus 0.5em minus 0.4em\relax PMLR, 2021, pp. 8748--8763.

\bibitem{li2023vihope}
H.~Li, S.~Dikhale, S.~Iba, and N.~Jamali, ``Vihope: Visuotactile in-hand object 6d pose estimation with shape completion,'' \emph{IEEE Robotics and Automation Letters}, 2023.

\bibitem{li2022discovering}
Y.-L. Li, H.~Fan, Z.~Qiu, Y.~Dou, L.~Xu, H.-S. Fang, P.~Guo, H.~Su, D.~Wang, W.~Wu \emph{et~al.}, ``Discovering a variety of objects in spatio-temporal human-object interactions,'' \emph{arXiv preprint arXiv:2211.07501}, 2022.

\bibitem{zhao2022dualafford}
Y.~Zhao, R.~Wu, Z.~Chen, Y.~Zhang, Q.~Fan, K.~Mo, and H.~Dong, ``Dualafford: Learning collaborative visual affordance for dual-gripper object manipulation,'' \emph{arXiv preprint arXiv:2207.01971}, 2022.

\bibitem{xue2023arraybot}
Z.~Xue, H.~Zhang, J.~Cheng, Z.~He, Y.~Ju, C.~Lin, G.~Zhang, and H.~Xu, ``Arraybot: Reinforcement learning for generalizable distributed manipulation through touch,'' \emph{arXiv preprint arXiv:2306.16857}, 2023.

\bibitem{di2024effectiveness}
N.~Di~Palo and E.~Johns, ``On the effectiveness of retrieval, alignment, and replay in manipulation,'' \emph{IEEE Robotics and Automation Letters}, 2024.

\end{thebibliography}
